\documentclass[sigconf]{acmart}
\AtBeginDocument{%
  \providecommand\BibTeX{{%
    \normalfont B\kern-0.5em{\scshape i\kern-0.25em b}\kern-0.8em\TeX}}}

\acmConference[]{}{}{}
\usepackage{caption}
\usepackage{subcaption}
\usepackage{graphicx}
\begin{document}

\title{Explain Variance of Prediction in Variational Time Series Models for Clinical Deterioration Prediction}

\author{Jiacheng Liu}
\email{liu00520@umn.edu}
\orcid{0000-0002-1449-5038}
\affiliation{%
  \institution{University of Minnesota, Twin Cities}
  \city{Minneapolis}
  \state{Minnesota}
  \country{USA}
}

\author{Jaideep Srivastava}
\email{Srivasta@umn.edu}
\orcid{0000-0000-0000-0000}
\affiliation{%
  \institution{University of Minnesota, Twin Cities}
  \city{Minneapolis}
  \state{Minnesota}
  \country{USA}
}

\renewcommand{\shortauthors}{Liu and Srivastava}

\begin{abstract}
Missingness and measurement frequency are two sides of the same coin. How frequent should we measure clinical variables and conduct laboratory tests?  It depends on many factors such as the stability of patient conditions, diagnostic process, treatment plan and measurement costs. The utility of measurements varies disease by disease, patient by patient. In this study we propose a novel view of clinical variable measurement frequency from a predictive modeling perspective, namely the measurements of clinical variables reduce uncertainty in model predictions. To achieve this goal, we propose variance SHAP with variational time series models, an application of Shapley Additive Expanation(SHAP) algorithm to attribute epistemic prediction uncertainty. The prediction variance is estimated by sampling the conditional hidden space in variational models and can be approximated deterministically by delta's method. This approach works with variational time series models such as variational recurrent neural networks and variational transformers. Since SHAP values are additive, the variance SHAP of binary data imputation masks can be directly interpreted as the contribution to prediction variance by measurements. We tested our ideas on a public ICU dataset with deterioration prediction task and study the relation between variance SHAP and measurement time intervals.
\end{abstract}

% \BEGIN{CCSXML}
% <CCS2012>
%  <CONCEPT>
%   <CONCEPT_ID>00000000.0000000.0000000</CONCEPT_ID>
%   <CONCEPT_DESC>DO NOT USE THIS CODE, GENERATE THE CORRECT TERMS FOR YOUR PAPER</CONCEPT_DESC>
%   <CONCEPT_SIGNIFICANCE>500</CONCEPT_SIGNIFICANCE>
%  </CONCEPT>
%  <CONCEPT>
%   <CONCEPT_ID>00000000.00000000.00000000</CONCEPT_ID>
%   <CONCEPT_DESC>DO NOT USE THIS CODE, GENERATE THE CORRECT TERMS FOR YOUR PAPER</CONCEPT_DESC>
%   <CONCEPT_SIGNIFICANCE>300</CONCEPT_SIGNIFICANCE>
%  </CONCEPT>
%  <CONCEPT>
%   <CONCEPT_ID>00000000.00000000.00000000</CONCEPT_ID>
%   <CONCEPT_DESC>DO NOT USE THIS CODE, GENERATE THE CORRECT TERMS FOR YOUR PAPER</CONCEPT_DESC>
%   <CONCEPT_SIGNIFICANCE>100</CONCEPT_SIGNIFICANCE>
%  </CONCEPT>
%  <CONCEPT>
%   <CONCEPT_ID>00000000.00000000.00000000</CONCEPT_ID>
%   <CONCEPT_DESC>DO NOT USE THIS CODE, GENERATE THE CORRECT TERMS FOR YOUR PAPER</CONCEPT_DESC>
%   <CONCEPT_SIGNIFICANCE>100</CONCEPT_SIGNIFICANCE>
%  </CONCEPT>
% </CCS2012>
% \END{CCSXML}




\maketitle

\section{Introduction}
%background
%explainability
\subsection{Background}
Come with the mighty power of deep learning models, is the disagreeable feature of being black boxes. As deep learning models are gaining huge success in many areas, the problem of low explainability and interpretability become more and more urgent, especially in the healthcare domain, where the risk-benefit trade-offs are constantly being considered against patients and the cost of an decision error could be disastrous. Therefore, to make real impacts on healthcare decisions, it is necessary and fundamental to understand and explain how and why a black-box deep learning model produce certain predictions. Following this thought, we also argue that the explanations of model predictions must be local, in a sense that it is relevant to individual patients, instead of only being valid on a large population level. Besides locality, in order to suggest actionable insights from the outputs of models, the explanations must be human-consumable\cite{Di_Martino_2022} and land at specific, concrete domain concepts. Otherwise, we will soon find ourselves in an endless game of explain an explanation. Last but not least, explanations must be validated clinically as the gap between accurate model predictions and desirable outcomes(e.g., improved prognosis, reduced cost) are non trivial.

Researchers have made enormous amount of efforts to make black boxes transparent. Both model specific and model agnostic methods have been developed to tackle the challenge of explaining the outputs of deep learning models. Among them, the game theoretic approach of SHAP\cite{huang2022encoding,nie2022time} (\textbf{SH}apley \textbf{A}dditive ex\textbf{P}lanations) stands out and become one of the most popular method. From the analyses in the first paragraph, it is easy to see why this is the case. To simply put, SHAP is basically a localized version of Shapley value. What's more, the additive and linearity also fit the needs of healthcare application perfectly. 

\subsection{Motivations}
Explanations about the predictive model output alone may not be enough. To gain a complete view of predictions, we need to understand its variability. For future event predictions, such as patient deterioration prediction, it is also desirable to understand how soon the event will happen. Traditionally, these tasks can be done by training separate models against different targets. However, this approach is at the risk of inconsistent explanation and predictions among different models, thus may be difficult for human to understand. This may cause a problem when trying to translate explanation results to actions, because the causal relation between different tasks are implicit. Therefore, the only valid solution is to use one model for all tasks. Multitask learning is the most appreciated approach in this case. Alternatively, and recently, generative models are also naturally capable of performing multitasks. We propose to use variational generative models to predict patient deterioration, meanwhile, desirable quantities like prediction variability, and acuity of disease(how fast the patient is deteriorating?) can be derived from the model. 

\subsection{Proposed Method}
%archi
\begin{figure*}[t]
  \centering
  \includegraphics[width=14cm]{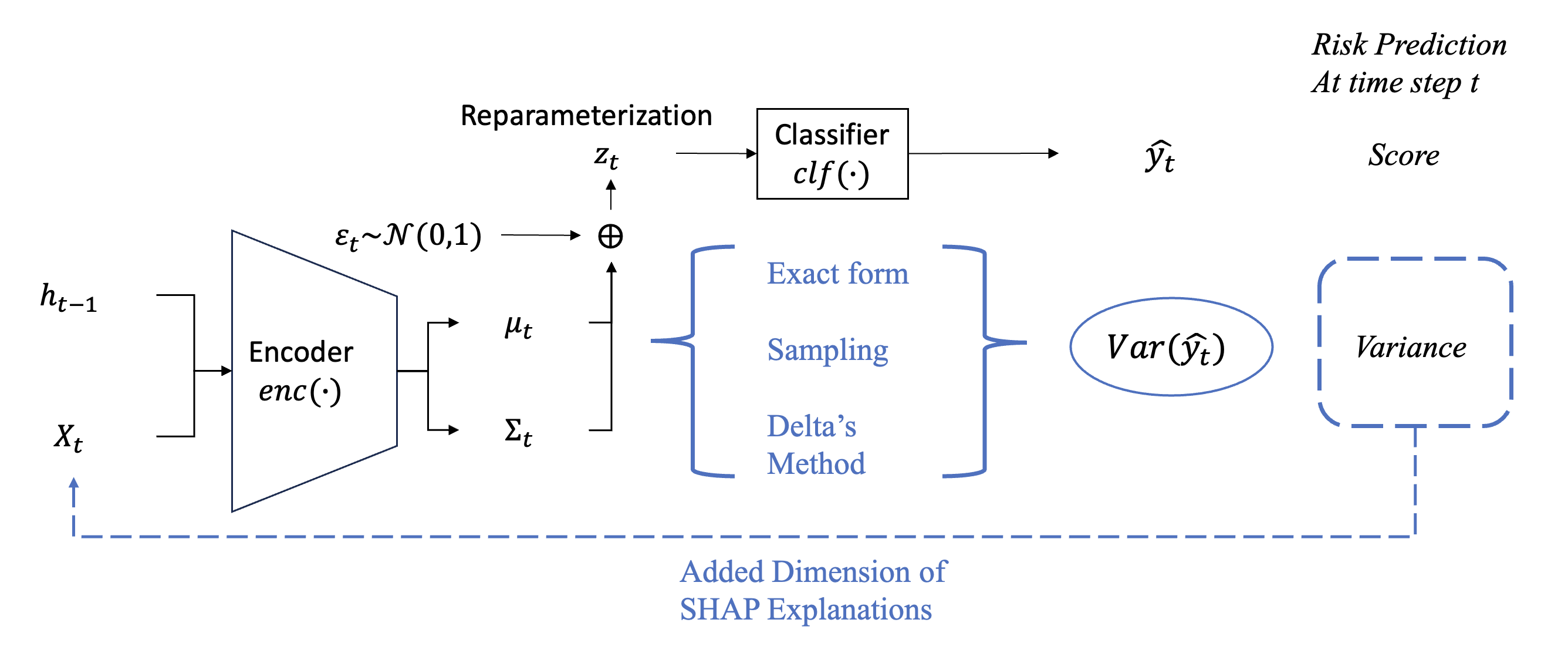}
  \caption{Proposed architecture for explaining prediction variance. Contributions in \textcolor{blue}{blue}. The prediction variance can be deterministically approximated from the posterior distribution of hidden states via Delta's method, which is essentially a Taylor expansion. }
  \label{fig:archi}
\end{figure*}
In this paper, we aim to exploit the locality and additivity of SHAP values, and expand model predictions to prediction variability, with the help of variational models and a bit of stochastic calculus. Variational inference methods are powerful generative models which approximate the posterior distribution of assumed hidden, unobserved variables. While in variational models, the hidden states are represented by random variables, we construct explicit and deterministic games of prediction variance, so that the SHAP values can be back propagated to input clinical variables. Further, we argue that additive SHAP values, propagated to carefully handcrafted features, can potentially translate to real actions. We provide such an example in the experiment section, where we discover that the frequency of clinical variable collections are highly correlated prediction variability. The idea is best illustrated by figure\ref{fig:archi}. A variational time series model will be trained for deterioration prediction. At every time step, SHAP value explanations of predicted risk score, prediction variance are generated along with the model output.

%Contributions
Our contributions are two-fold.
\begin{itemize}
    \item Variance SHAP. A deterministic game of variance is constructed such that the total prediction variance is explicitly calculated. Then, SHAP values are used to explain the contribution.

    \item Connection between variance SHAP and frequency of measurements. We further study the relation between variance SHAP and the frequency of measurements of clinical variables. Though most variables behave as expected, i.e. the longer interval between two measurements, the more prediction variance there will be. We discovered that there also exit abnormal patterns of this relation, namely the contribution to prediction variance decreases as the interval between two non-missing values increases.
\end{itemize}

%structure
The paper is organized as the follows. In section 2, we discuss the related work about estimating prediction uncertainty, SHAP method and its applications in healthcare. In section, we introduce preliminary concepts about variational time series models(the model) and patient deterioration prediction(the task). We first illustrate how the variance SHAP is different from SHAP of prediction and the variance of gradient. The using MNIST dataset\cite{deng2012mnist}. Finally, we conclude with the limitations and possible future directions of this study. Jupyter notebooks and codes for the public dataset are available on github\footnote{\url{ https://github.com/kanbei7/VarianceSHAP_for_VRNN}}.

\section{Related Work}
\subsection{Shapely Values and DeepSHAP}
Since the advent of SHAP\cite{lundberg2017unified,lundberg2020local}, it has been extensively applied to many areas, such as geology\cite{al2023novel}, finance\cite{godin2023risk,duan2022factorvae} and healthcare\cite{chen2022explaining,lundberg2018explainable}.
Recently, SHAP values have been applied to gaussian process\cite{chau2023explaining}. The authors have shown that the variance of SHAP value (which is necessary for inferences of gaussian process) is not the SHAP value of the variance game(our focus in this paper). SHAP has also been combined with variational auto encoders (VAE) to explain feature contributions\cite{olsen2022using,withnell2021xomivae}. We appreciate the beauty of SHAP values since it is model agnostic and flexible to most type of model outputs. To take a step further, if there exists a function calculating the variance of the prediction, SHAP methods can be applied as well!
 
We also note that there are plenty of approaches other SHAP values, especially for time series data\cite{lim2021temporal,castro2023time,crone2010feature,taieb2015bias}. However, as argued in the previous section, we claim that SHAP is the best fit for our case. Besides, studies have shown several drawbacks of SHAP, noticeably with entangled time series features\cite{ismail2020benchmarking}. But as we shall see in later sections, with the power of variational time series models, this defect is mitigated by the inference of independent hidden state variables and thus not a major concern to our topic.

\subsection{Prediction Uncertainty and Variance}
The predicted probability score by machine learning models contains two sources of uncertainty, aleatoric and epistemic uncertainty\cite{hullermeier2021aleatoric,swiler2009epistemic}. While the latter one comes from the uncertainty about model parameters, aleatoric originates from data and unobserved factors. With aleatoric uncertainty and variance of prediction scores in focus, Bayesian methods\cite{gal2016dropout,kendall2017uncertainties,neal2012bayesian} and variational inference methods\cite{kingma2013auto,rezende2014stochastic} become natural choices for estimating prediction variance, with the assumption that hidden state random variables approximate the posterior distribution of inputs. This fundamental assumption should hold true for every variational generative model, for models to be fully effective. In this study we will focus on how to explain the contributions of input clinical features to prediction variance.

\subsection{Explainable AI for time series in healthcare}
In an application area like healthcare, explainability and interpretability are a crucial features to build trustworthy machine learning and AI applications.
\cite{Di_Martino_2022} has a nice summary of recent development of recent Explainable Artificial Intelligence(XAI) in healthcare. SHAP values is the dominant approach in recent studies about explainable healthcare machine learning models\cite{Ang2021-ws,ukil2022less},. While few studies have focused on explaining deep learning models\cite{withnell2021xomivae,Oviedo2019-xk}, fewer studies have focused on explaining time series in healthcare\cite{lundberg2018explainable,ivaturi2021comprehensive}. The most related work is that \cite{withnell2021xomivae} propose to use variational auto encoders to study multi-omics data for cancer diagnostic. However, to the best of our knowledge, we are the first to study contributions of prediction variance and the first to attempt to study the relation between variance SHAP and the frequency of variable measurements.

\section{Preliminaries}
In this section, we introduce preliminary concepts of the problem as well as notations and abbreviations. For simplicity, all notations omit the batch dimension. 
\subsection{Variational Time Series Models}
As the name suggests, variational time series models combine variational inference with recurrent structures for time series. We refer the term \textit{``variational time series models''} to any time series models \textbf{(i)} of which its hidden states are represented by a set of parameters of some probability distributions and \textbf{(ii)} its hidden states are updated by some recurrence mechanism, e.g. recurrent gated unit. A wide range of models fall into this category, such as variational recurrent neural networks(VRNN)\cite{chung2015recurrent,purushotham2016variational}, stochastic recurrent neural networks(SRNN)\cite{bayer2014learning,fraccaro2016sequential}, and variational transformers(VTrans)\cite{shamsolmoali2023vtae,lin2020variational}.

\subsubsection{Framework}
Formally, we denote the input time series as $\textbf{x} = (x_1, x_2, ... , x_n)$ where $n$ is the length of the sequence, and subscription $t$ is a dummy variable for time step. Each $x_t \in \mathbb{R}^{d}$, with $d$ being the number of features. The deterministic hidden states from the recurrent model is marked by $h_t$, while the random variable $z_t$ is drawn from a set of distributions parameterized by $\mu_t$ and $\Sigma_t$. $\theta_t$ is a short hand of the combination of distribution parameters $\mu_t$ and $\Sigma_t$. $\theta_{t,prior}$ stands for the parameters of prior distribution. $\hat{y_t}$ and $y_{t}$ denote the predicted score and the ground truth respectively. Though the main task can be of various kind such as classification, prediction or regression, we use $CLF(\cdot)$ for the main task network. Similarly, 
$RNN(\cdot)$ for recurrence unit, but really it can be any recurrence mechanism like Long-Short Term Memory\cite{hochreiter1997long}, Gated Recurrent Unit\cite{Chung2014-cn} or Transformers\cite{huang2022encoding,nie2022time}. The naming of other components are straightforward, $ENC(\cdot)$ for encoders, $DEC(\cdot)$ for decoders and $PRIOR(\cdot)$ for prior network. 
For parameter of distributions,  $\mu_t,\Sigma_t  \in \mathbb{R}^{z_dim}$. For hidden states, $h_t \in \mathbb{R}^{h_dim}, z_t  \in \mathbb{R}^{z_dim}$.

\begin{equation}
\begin{split}
      \theta_{t,prior} &= PRIOR(h_{t-1})\\
  \theta_{t} &= ENC(h_{t-1}, x_t), \theta_{t} = \mu_t, \Sigma_t
\end{split}
\end{equation}

\begin{equation}
    z_t = \mu_t + \Sigma_t\epsilon_t, \epsilon_t \sim N(0,1) 
\end{equation}

\begin{equation}
    \hat{y_t} = CLF(z_t)
\end{equation}

\begin{equation}
    h_t = RNN(h_{t-1}, x_t, z_t)
\end{equation}

Reconstructed $\hat{x_t}$ is given by the decoder. Based on the result of $\sigma$-Variational Auto Encoders\cite{rybkin2021simple}, it suffices to just out put $\hat{x_t}$ since the negative log likelihood loss is analytically determined by $\hat{x_t}$ itself. Therefore, there is no need for decoders to produce another set of parameters.

\begin{equation}
    \hat{x_t} = DEC(h_{t}, z_t)
\end{equation}

\begin{figure}[t]
  \centering
  \includegraphics[width=\linewidth]{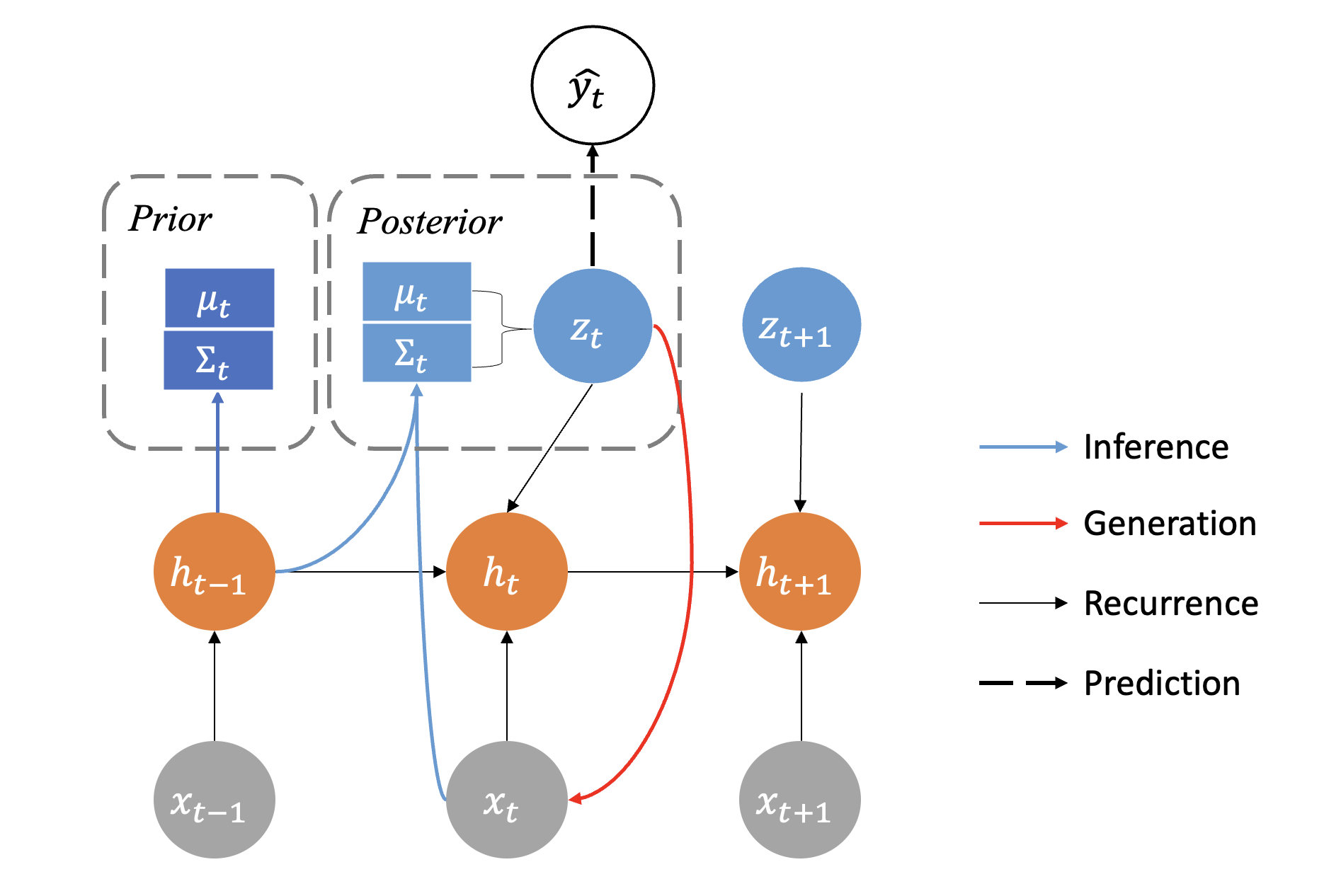}
  \caption{Variational Recurrent Models}
  \label{fig:vrnn}
\end{figure}

Figure \ref{fig:vrnn} shows a graphic representation of a typical variational time series model structure. Suppose the hidden space at time step $t$ consists of several independently normally distributed variables $z_t$, parameterized by mean vector $\mu_t$ and a diagonal covariance matrix $\Sigma_t$. Upon every time step forward,  $\mu_t$ and $\Sigma_t$ are inferred by the encoder network from current input $x_t$ and previous recurrent hidden state $h_{t-1}$. Next, the hidden state vector is drawn from the distribution based on inferred parameters via reparameterization tricks\cite{rezende2014stochastic,kingma2013auto}. The task network then uses $z_t$ to make predictions or classifications. As for the recurrent mechanism, both $z_t$ and $x_t$ serve as the inputs to the recurrence unit. In this way, the model allows for certain degrees of stochasticity or transition between hidden states. 

\subsubsection{Training}
%VAE ELBO
Since the integral of joint probability $p(x,z)$ over the entire hidden space, \[ \int_{\textbf{Z}}  p(x,z')\,dz' \], is intractable, variational are trained on evidence lower bound (ELBo), given by.
\begin{equation}
    \begin{split}
        ELBo =& ln p_{\theta}(x|z,h) \\
        &- KLdivergence[p_{\theta}(z|x,h)||p_{\theta_{prior}}(h)] \\
    \end{split}
\end{equation}
Maximizing ElBo is equivalent to minimizing $\mathcal{L}_{kld} + \mathcal{L}_{NLL}$, defined below. Subscription $t$ applies to all variables above, hence is omitted.

The training of variational time series models shares similarities with variational auto encoders (VAE)\cite{rezende2014stochastic,kingma2013auto,sohn2015learning}: a Kullback–Leibler divergence between posterior $\theta_{t}$ and prior $\theta_{t,prior}$ and a regularization loss on $\theta_{t}$ . We adopt the approach in $\sigma$-Variational Auto Encoders\cite{rybkin2021simple} such that:
\begin{equation}
\begin{split}
    \mathcal{L}_{kld} & = KLdivergence[p_{\theta_{t}}(z_t|x_t,h_t)||p_{\theta_{t,prior}}(h_t)] \\
    & p_{\theta_{t}}(z_t|x_t,h_t) \sim N(\mu_{t},\Sigma_{t})\\
   & p_{\theta_{t,prior}}(z_t|h_t) \sim N(\mu_{t,prior},\Sigma_{t,prior})
\end{split}
\end{equation}

\begin{equation}
\begin{split}
    \mathcal{L}_{\sigma-NLL} &=NLL(\hat{x_t},x_t,log(\sigma)) \\
    log(\sigma) &= log(\sqrt{(\hat{x_t}-x_t)^2})
\end{split}
\end{equation}

Notice that there are different ways of choosing a prior, depending on the specific problem settings. These two losses are essential components for variational inference.

%prediction target
Additionally, we have the prediction or classification loss from the main task network and the reconstruction loss to train recurrent networks. $MSE$ denotes the mean squared error.
\begin{equation}
\begin{split}
    \mathcal{L}_{clf} &= Cross-Entropy(\hat{y_t},y_{t}) \\
    \mathcal{L}_{recon} &= MSE(\hat{x_t},x_t) 
    \end{split}
\end{equation}
Therefore the total loss per time step is given by the following equation.
\begin{equation}
\begin{split}
        \mathcal{L}_{total} =& \mathcal{L}_{kld} + \mathcal{L}_{\sigma-NLL} + \mathcal{L}_{clf} \\
        & +  \mathcal{L}_{recon} + \lambda\mathcal{L}_{reg}
\end{split}
\end{equation}
%ts
Depending on whether a complete sequence can be seen at the time of inference, the loss may be average over all the time step. Also, regularization terms $\mathcal{L}_{reg}$ may be appended to stabilize and accelerate training. $\lambda$ controls the strength of regularization.

\subsection{Patient Deterioration Prediction}
%task illustration
\begin{figure}[t]
  \centering
  \includegraphics[width=\linewidth]{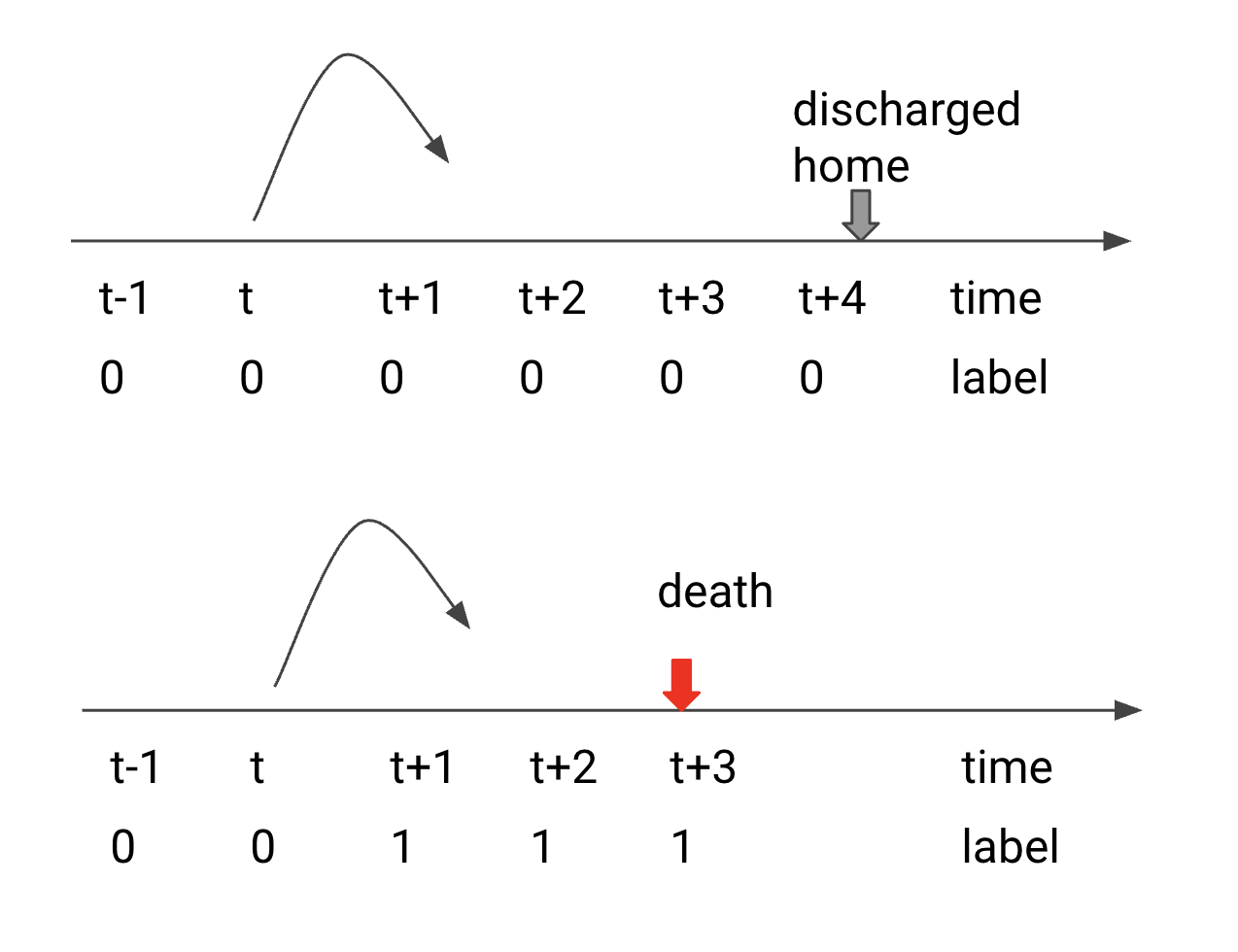}
  \caption{Patient Deterioration Prediction Task\cite{Liu2022-nb, harutyunyan2019multitask}}
  \label{fig:label}
\end{figure}
Clinical deterioration (also known as “clinical decompensation”) refers to the process during which the patient’s condition evolves towards undesirable outcomes. Depending on the context, its meaning varies. In the emergency room, the practice of predicting deterioration is known as “triaging” . Namely, to stratify patients based on their risk of deterioration, so that patients with immediate risk will be prioritized. For patients admitted to the Intensive Care Unit (ICU), physicians are concerned with unexpected worsening of the disease and risk of mortality. For patients in general wards, clinical deterioration usually results in critical events such as transfer to the ICU or cardiopulmonary arrest. The hope is that early predictions of onsets of clinical deterioration will eventually bring benefits to all stakeholders including patients, physicians, and insurance companies. In the scope of this paper, our models predict ICU transfers for general ward patients and risk of mortality for ICU patients. Figure \ref{fig:label} illustrates the n-step ahead risk of mortality predictions for ICU patients. The prediction is made at every time step.

\section{The Variance Game of Variational Models}
%motivating examples
%VAE
Although the hidden variables are modeled by parameterized distributions, the variance game(the game of attribute variance)\cite{galeotti2021comparison,colini2018variance}, is actually deterministic, because we can explictly and deterministically calculate the variance of $\hat{y}=clf(z)$ given $z\sim N(\mu, \Sigma)$. The only thing we need to do is to wrap the original model so that the prediction variance become the wrapped model's output. Then, SHAP method can be applied. We use $v(\cdot)$ to denote the value of a game. We disregard sampling methods for its computational cost.

\subsection{Logits of Sigmoid}
The simplest case is to have logits as prediction model output. In this case, we can have an exact form of variance, thanks to the linearity of variance\cite{galeotti2021comparison}. The value of the variance game is just the linear combination of all variance. Assume $w$ to be weights of the last linear layer. Then,
\begin{equation}
    v(logits)=\Sigma w_{i}^2\sigma_{i}^2
\end{equation}

\subsection{Delta's Method}
For more complicated $clf$ functions, we resolve the problem by using Delta's method\cite{oehlert1992note,hong2018numerical}. To estimate $var[f(z)]$, where $z \sim N(\mu, \sigma)$, notice that 

\begin{equation}
    f(z) \approx f(\mu) + (z-\mu)f'(\mu)
\end{equation}

Therefore, the variance can be estimated by,

\begin{equation}
\begin{split}
    var[f(z)] = & var[f(z-\mu)]\cdot[f'(\mu)]^2 \\
    & \sigma^2\cdot[f'(\mu)]^2
\end{split}
\end{equation}

%drawbacks of the sampling method
\subsection{MNIST as an illustration}

\begin{figure*}[h]
\centering
  \includegraphics[width=14cm]{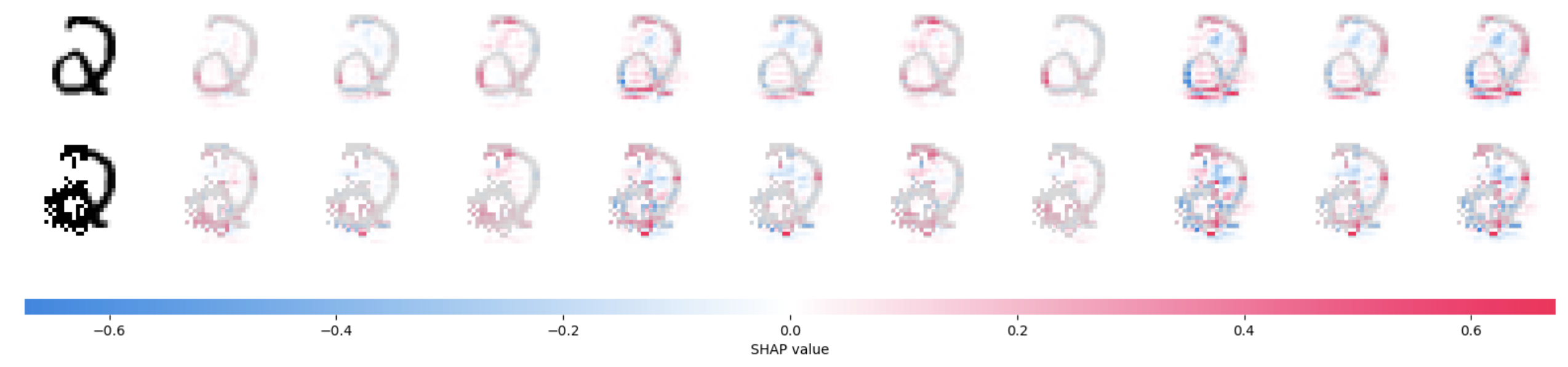}
  \caption{Prediction SHAP value.}
  \label{fig:SHAPpred}
\end{figure*}

\begin{figure*}[h]
  \includegraphics[width=14cm]{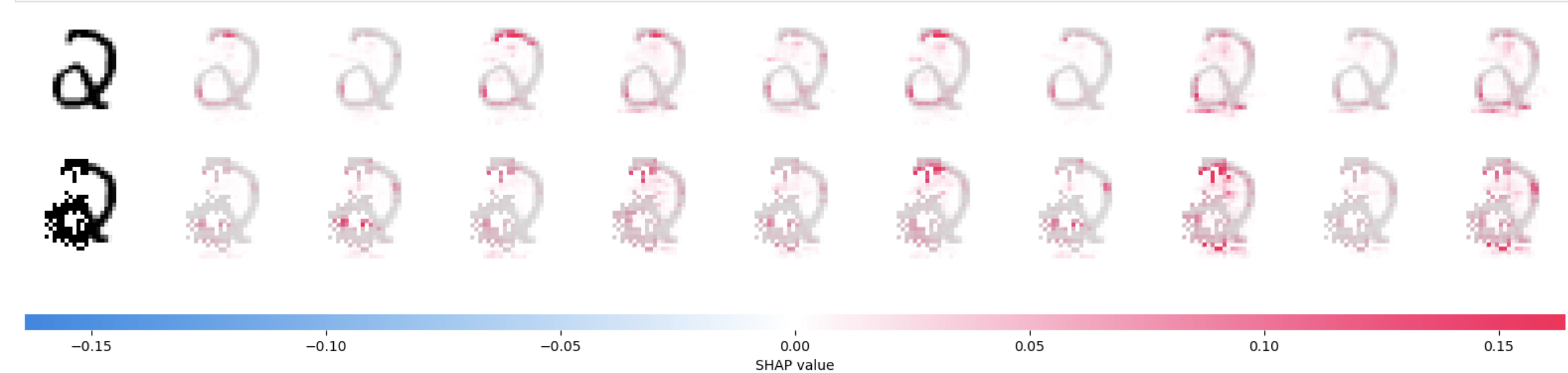}
  \caption{Variance of the prediction SHAP value}
  \label{fig:SHAPvar}
\end{figure*}

\begin{figure*}[h]

  \includegraphics[width=13.7cm]{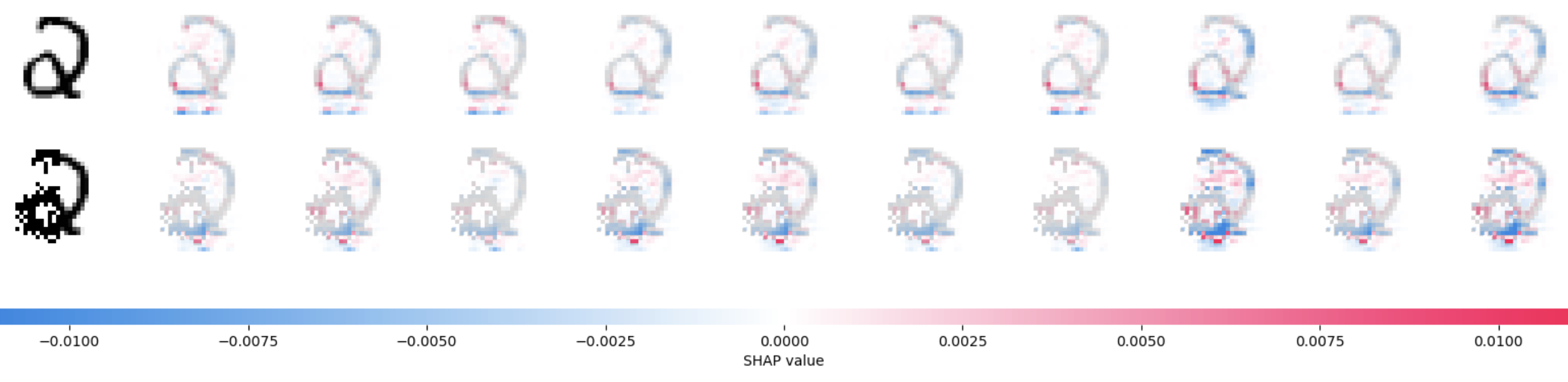}
  \caption{SHAP value of prediction variance attribution.}
  \label{fig:varSHAP}
\end{figure*}
  
%\caption{Comparision of (a)SHAP values of predicted class probabilities, (b) Variance of the prediction SHAP, (c)(proposed)SHAP value of prediction variance attribution}
We verify our proposed method by training a unidirectional variational recurrent neural network on MNIST. The training took 10 epochs and achieved an accuracy of 98\%. Details are provided in the Jupyter notebook.Figure\ref{fig:SHAPpred},\ref{fig:SHAPvar},\ref{fig:varSHAP} compare (a)SHAP values of predicted class probabilities, (b) Variance of the prediction SHAP, (c)(proposed)SHAP value of prediction variance attribution. The first thing we noticed is that as expected, the attribution of variance and prediction do not coincide. Namely the model can be very confident on the prediction with a high predicted probability score(high score with low variance), or the model can be sure that the input instance is not one of the target class (low score with low variance). Vice versa, there could be the case where high score and high variance co-exist. Additionally, as a note to the discussion in related work, we didn't observe SHAP attributions as chaotic as reported in \cite{ismail2020benchmarking} (on the same MNIST dataset). Therefore, we did not apply the normalization technique of temporal saliency mapping(TSR).
\section{Experiments}
\subsection{MIMIC-IV}
Medical Information Mart for Intensive Care, or MIMIC-
IV in short, is a large, open-source, deidentified database of
hospitalized patients\cite{johnson2020mimic,johnson2023mimic,goldberger2000physiobank}. It contains clinical notes, ECG
data, time series of vital signs, laboratory test results and
assessment scores. In this study, we use MIMIC4
data for ICU patients which contains about 60,000 ICU stays after data cleaning. The data is aggregated to hourly level time series of varying lengths. The median length of stay is 61 hours, while the mortality rate is 7.8\%. We train variational recurrent models to predict the risk of mortality in the next 48 hours. All variables are normalized and sanity checked(for example, heart rate can not be negative, saturation of oxygen must be within 0 to 100, all temperatures share the same units, etc.). The process left 176 time series variables. We pick 10 of them after extensive feature selections. In addition, a mask is associated with each variable indicating whether the value is missing and imputed or actually measured. Log base 24 is also applied to time intervals between measurements. Therefore, there is a total of 30 features per time step. The dataset is split to training, validation and test set, controlling both mortality rate length of stay.

\subsection{Variance attribution and frequency of measurements}
\begin{figure*}[h]

  \includegraphics[width=14cm]{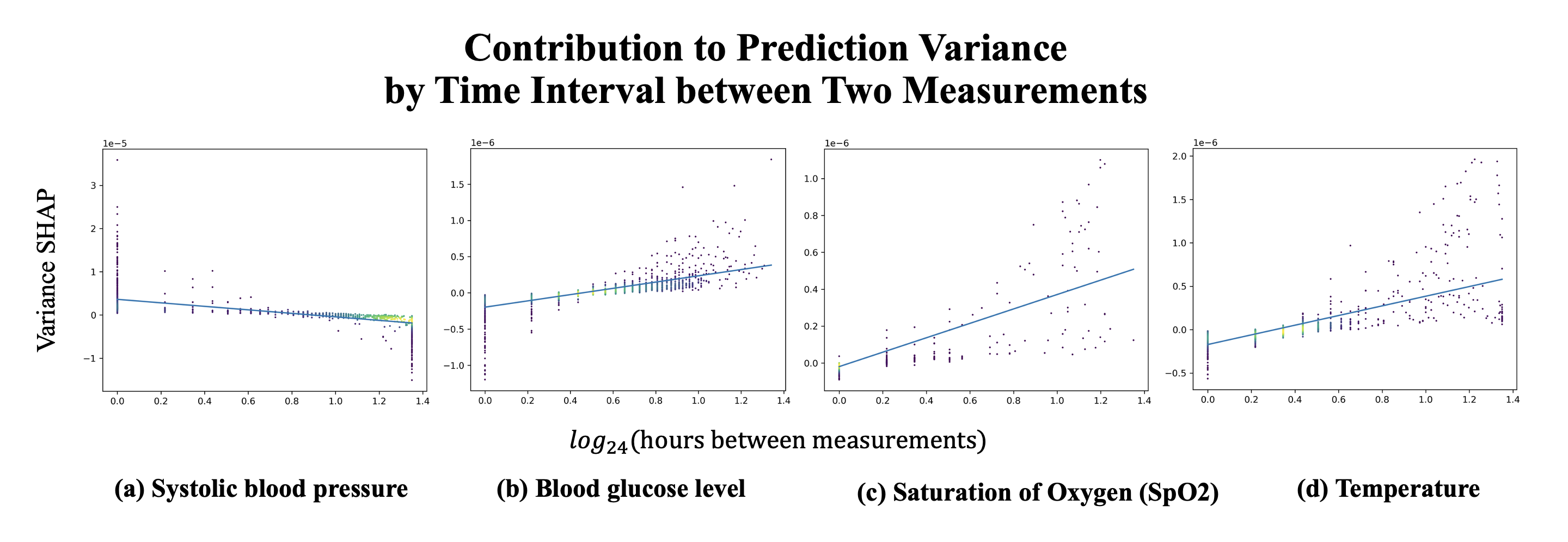}
  \caption{Relation between variance SHAP and time interval between measurements}
  \label{fig:normaltime}
\end{figure*}

\begin{figure*}[h]

  \includegraphics[width=14cm]{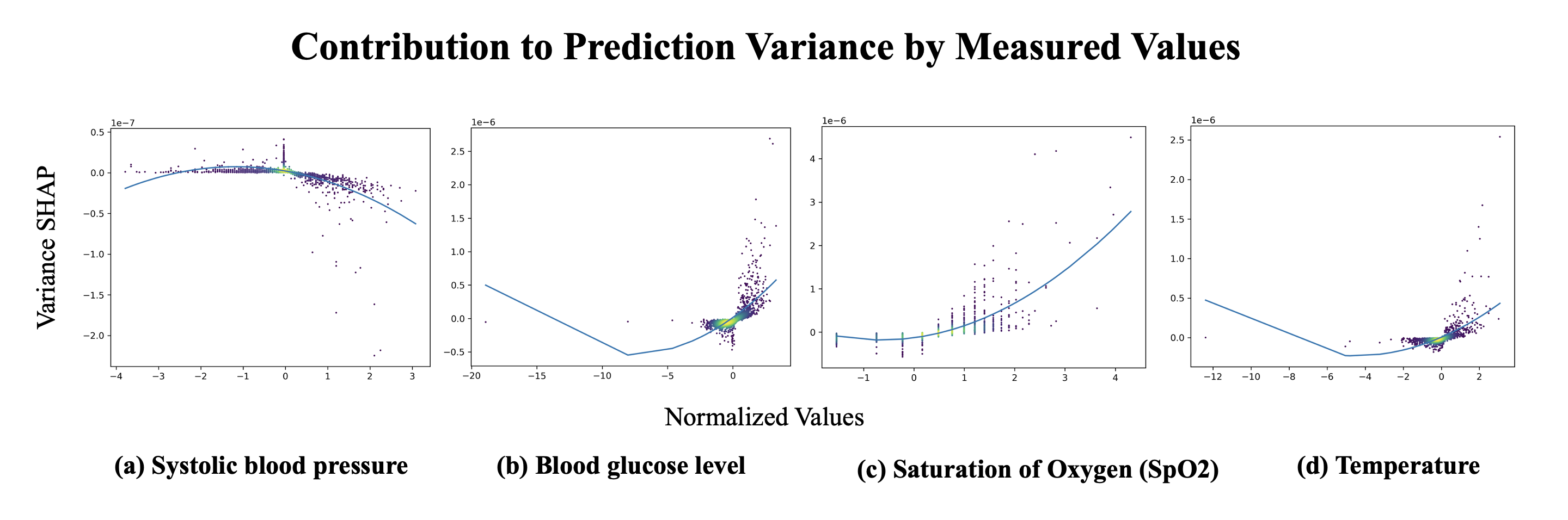}
  \caption{Relation between variance SHAP and the measured value.}
  \label{fig:normalvalue}
\end{figure*}

\begin{figure*}[h]
  \centering
  \includegraphics[width=14cm]{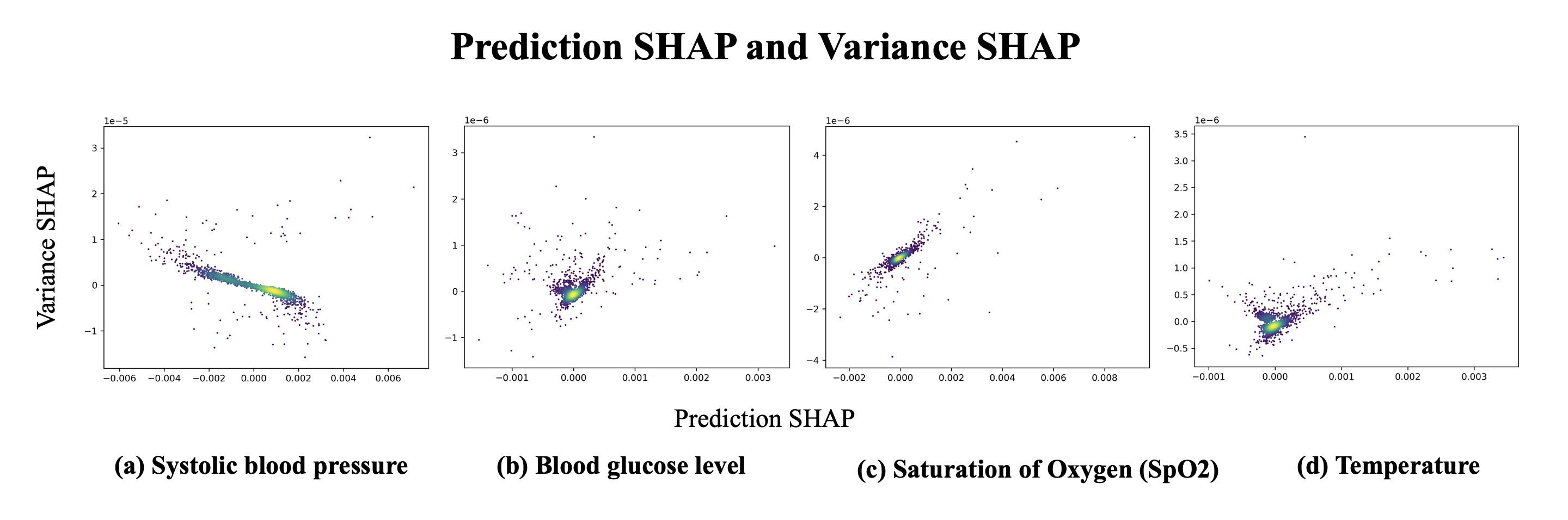}
  \caption{Relation between the contribution to predicted mortality risk (prediction SHAP) and contribution to prediction variance (variance SHAP)}
  \label{fig:predvar}
\end{figure*}
To model the frequency of measurements, we simply handcraft features that represents the log time interval from the last valid measurements. In this way, by checking the variance SHAP contribution of these variables, we attempt to answer questions: how frequent we should measure a specific clinical variable? Are there excessive and unnecessary frequent measurements which can be avoided? Due to the size of the data, we randomly sample 6,000 patients from training set as background. Still, it took about 12 hours on a machine equipped with RTX3090 to check the variance contribution at every time step. We select the variance contribution at 72 hours as an example. The figure below shows a typical pattern that we would expect. Basically the contribution to prediction variance would increase as the time interval increases.

\subsection{Abnormal patterns}
However, we have also observed abnormal patterns for blood pressure measurements. Notice that for blood pressure, the zero score no longer falls into normal reference range. The average blood pressure of MIMIC-IV ICU population is well above the reference range. Therefore, the interpretation may be different.
\begin{figure*}[h]

  \includegraphics[width=13.7cm]{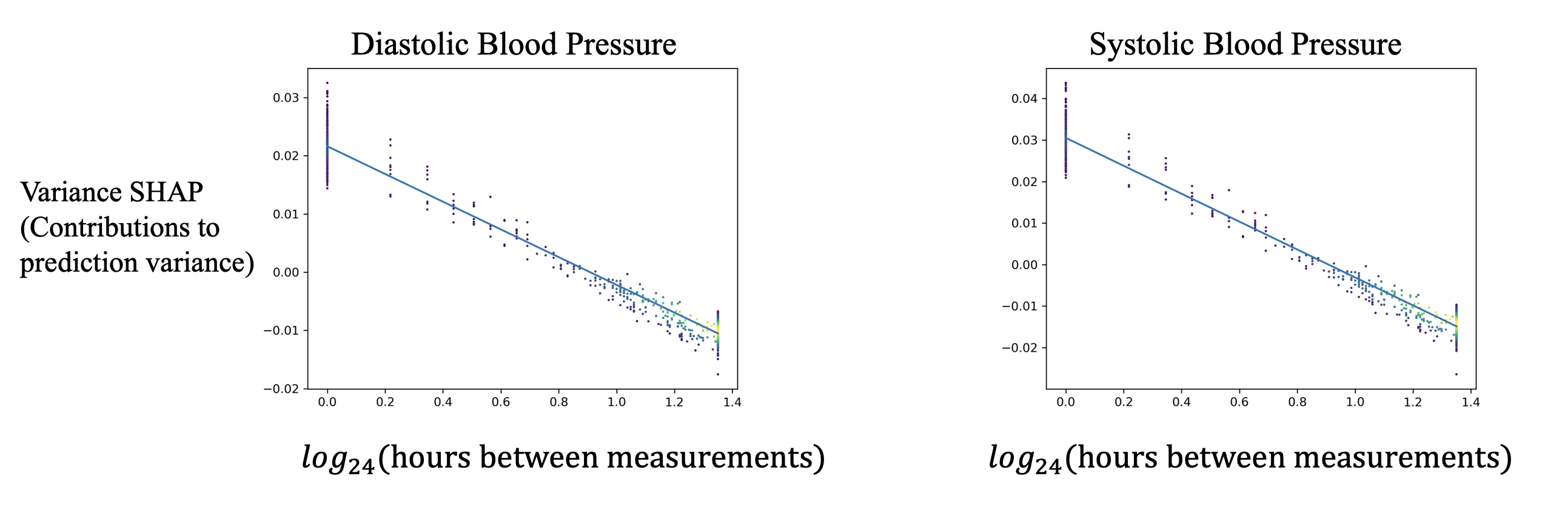}
  \caption{Unexpected patterns of blood pressure. There exist negative correlation predicction variance and between time interval}
  \label{fig:varSHAP}
\end{figure*}

\begin{figure*}[h]

  \includegraphics[width=13.7cm]{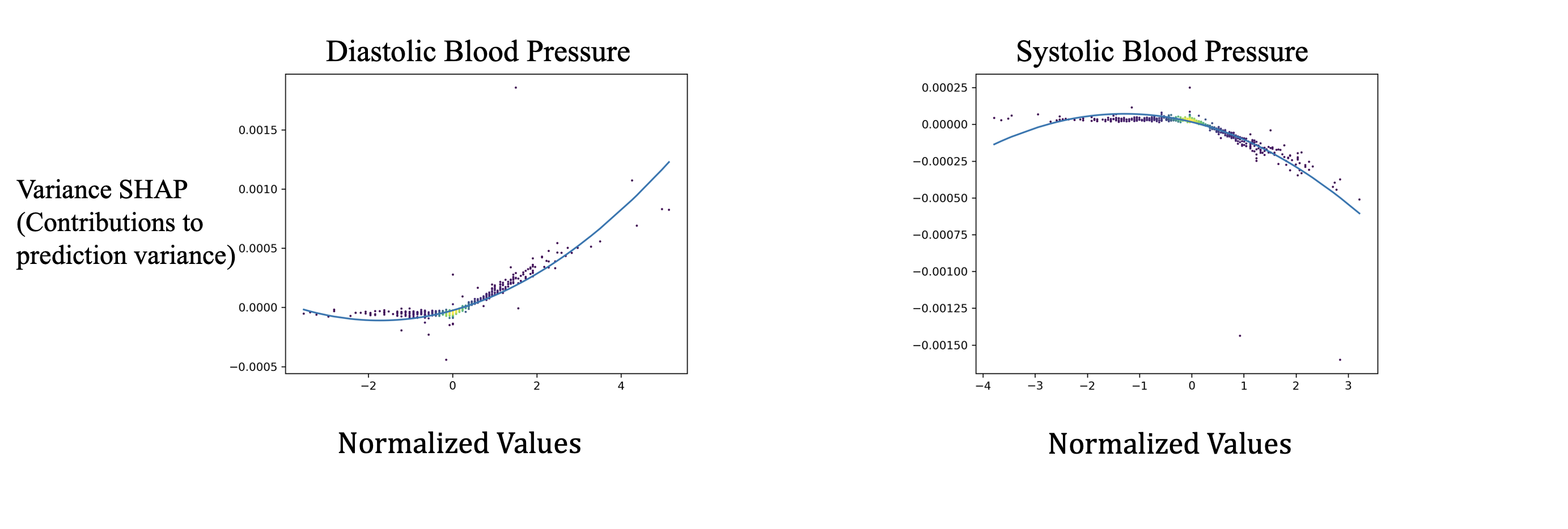}
  \caption{Unexpected patterns of blood pressure.}
  \label{fig:varSHAP}
\end{figure*}

\subsection{Avoidable and Missing Measurements}
For measured variables which contribute (either positively or negatively) little to both predicted risk score and prediction variance, we define these measurements as potentially ``\textit{avoidable measurements}''. For variables of which its missingness contribute significantly to prediction variance, we mark these measurements as potentially ``\textit{should-have measurements}''. The table below shows part of the statistics of sampled test dataset at 72 hours length of stay.
We expect to have a better understanding of when more variables and lab test results are included in the prediction and variance explanation model.

\begin{table*}[h]
\centering
\begin{tabular}{lllllll}
\toprule
\multicolumn{7}{c}{\textbf{Avoidable and Should-have Measurements for ICU patients}}                            \\
\hline
\midrule
\textbf{variable}       & \#avoidable & \#existing & \%      & \#should-have & \#missing & \%      \\
\textbf{Systolic\_BP}   & 0           & 523                     & 0.00\%  & 709           & 1477      & 48.00\% \\
\textbf{Glucose} & 161         & 366                     & 43.99\% & 640           & 1634      & 39.17\% \\
\textbf{Spo2}           & 711         & 1811                    & 39.26\% & 140           & 189       & 74.07\% \\
\textbf{Temperature}    & 133         & 497                     & 26.76\% & 542           & 1503      & 36.06\% \\
\bottomrule
\end{tabular}
    \caption{Avoidable and Should-have Measurements of Sampled patients at time = 72 hours}
    \label{tab:sumstat}
\end{table*}

\section{Discussion}

%TS SHAP problems
We have acknowledged that there works pointing to the problems of SHAP values. For example, \cite{ismail2020benchmarking} points out that SHAP does not work well with time series models. However, we didn't observe SHAP attributions as chaotic as reported in \cite{ismail2020benchmarking} (on the same MNIST dataset). Therefore, we did not apply the normalization technique of temporal saliency mapping(TSR).
%Additivity

%interpretation of variance-frequency
There are several limitations of this work. First, we are not able to find a fair baseline or ground truth model to compare and validate the efficacy. Second, since the delta's method is an approximation of prediction variance, to gain more accurate estimations, it may be desirable to further expand the Taylor series to include the second order derivatives.Last but not least, clinical validations and further investigations are needed for

For future work, it is intriguing to study the reason behind abnormal patterns. Besides, since SHAP value measure the difference between local feature contribution to expected output, looking into the absolute value of variance contribution may also be helpful in clinical settings. Another potential application is to search for potentially avoidable order of lab tests without compromising the quality of care. Thus the cost can be reduced. This would be useful especially for pediatric care, where frequent blood draws may bring more harm than benefits.

\section{Conclusions}
In this paper, we  proposed a deterministic game of prediction variance so that the contribution of individual input features can be calculated by the SHAP algorithm. What's more, we studied the connection between variance SHAP values and the  We further study the relation between variance SHAP and the frequency of measurements of clinical variables. Though most variables behave as expected, i.e. the longer interval between two measurements, the more prediction variance there will be. We discovered that there also exit abnormal patterns of this relation, namely the contribution to prediction variance decreases as the interval between two non-missing values increases. Finally, we discuss the limatations and possible future work.


%   \begin{acks}
% Identification of funding sources and other support, and thanks to
% individuals and groups that assisted in the research and the
% preparation of the work should be included in an acknowledgment
% section, which is placed just before the reference section in your
% document.
%   \end{acks}

\bibliographystyle{ACM-Reference-Format}
\bibliography{references}

\end{document}